%% file: 2018_cvpr.tex
\documentclass[10pt,twocolumn,letterpaper]{article}
\usepackage[table,dvipsnames]{xcolor}
\usepackage{tikz}  % For plots, must be the first include
\usepackage{cvpr}
\usepackage{times}
\usepackage{graphicx}
\usepackage{amsmath}
\usepackage{amssymb}
\usepackage{textcomp}
\usepackage{booktabs}
\usepackage{makecell}

\usepackage{subfiles}
\usepackage{pgfplots} % For plots
\usepackage{pgfplotstable} % For plots
\pgfplotsset{compat=newest} % For plots
\usetikzlibrary{plotmarks} % For plots
\usetikzlibrary{colorbrewer} % Nice colors
\usepackage{booktabs} % Nice tables
\usepackage{multirow}
\usepackage{etoolbox,siunitx} % Align decimal point
\robustify\bfseries
\robustify\itshape
\usepackage{paralist} % Nice itemize

\usepackage{flushend}

\input{00_Includes}

\usepackage[pagebackref=true,breaklinks=true,letterpaper=true,colorlinks,bookmarks=false]{hyperref}

\cvprfinalcopy % *** Uncomment this line for the final submission

\def\cvprPaperID{899} % *** Enter the CVPR Paper ID here

\ifcvprfinal\fi
\begin{document}

%%%%%%%%% TITLE
\title{Deep Extreme Cut: From Extreme Points to Object Segmentation}

\author{K.-K. Maninis\thanks{First two authors contributed equally} \quad
	S. Caelles$^{*}$\quad
	J. Pont-Tuset \quad
	L. Van Gool\\
	Computer Vision Lab, ETH Z\"urich, Switzerland
}

\maketitle
%\thispagestyle{empty}

%%%%%%%%% ABSTRACT
\begin{abstract}
This paper explores the use of extreme points in an object (left-most, right-most, top, bottom pixels) as input to obtain precise object segmentation for images and videos.
We do so by adding an extra channel to the image in the input of a convolutional neural network (CNN), which contains a Gaussian centered in each of the extreme points.
The CNN learns to transform this information into a segmentation of an object that matches those extreme points.

We demonstrate the usefulness of this approach for guided segmentation (grabcut-style), interactive segmentation, video object segmentation, and dense segmentation annotation. We show that we obtain the most precise results to date, also with less user input, in an extensive and varied selection of benchmarks and datasets.
All our models and code are publicly available on \url{http://www.vision.ee.ethz.ch/~cvlsegmentation/dextr/}.
\end{abstract}
\vspace{-5mm}

%%%%%%%%% BODY TEXT
\subfile{01_Introduction}

\subfile{02_Related}

\subfile{03_Method}

\subfile{04a_Ablation}

\subfile{04b_InstanceSeg}

\subfile{04c_Annotation}

\subfile{04d_VideoSeg}

\subfile{04e_Interactive}

\section{Conclusions}
We have presented \ours{}, a CNN architecture for semi-automatic segmentation that turns extreme clicking annotations into accurate object masks; by having the four extreme locations represented as a \textit{heatmap} extra input channel to the network.
The applicability of our method is illustrated in a series of experiments regarding semantic, instance, video, and interactive segmentation in five different datasets; obtaining state-of-the-art results in all scenarios.
\ours{} can also be used as an accurate and efficient mask annotation tool, reducing labeling costs by a factor of 10.

\paragraph*{Acknowledgments:} Research funded by the EU Framework Programme for Research and Innovation Horizon 2020 (Grant No. 645331, EurEyeCase), and the Swiss Commission for Technology and Innovation (CTI, Grant No. 19015.1 PFES-ES, NeGeVA). We gratefully acknowledge support by armasuisse and thank NVidia Corporation for donating the GPUs used in this project. We thank Dim P. Papadopoulos for sharing the resources of~\cite{Pap+17} with us and Vittorio Ferrari for constructive comments and discussions.

{\small
\bibliographystyle{ieee}
\bibliography{2018_cvpr}
}

\end{document}

%% file: 00_Includes.tex
\definecolor{olivegreen}{RGB}{0,170,0}
\definecolor{darkred}{RGB}{220,100,10}
\definecolor{tealblue}{RGB}{20,100,200}

\definecolor{rowblue}{RGB}{220,230,240}

\newcommand{\longours}{{Deep Extreme Cut}}
\newcommand{\ours}{{DEXTR}}

\newcommand{\showonerow}[7]{%
\setlength{\fboxsep}{0pt}
\resizebox{\textwidth}{!}{%
\fbox{\includegraphics[height=#7\linewidth]{ims/#6/#1.png}}
\hfill
\fbox{\includegraphics[height=#7\linewidth]{ims/#6/#2.png}}
\hfill
\fbox{\includegraphics[height=#7\linewidth]{ims/#6/#3.png}}
\hfill
\fbox{\includegraphics[height=#7\linewidth]{ims/#6/#4.png}}
\hfill
\fbox{\includegraphics[height=#7\linewidth]{ims/#6/#5.png}}
\vspace{2mm}
}}

\newcommand{\showonerowsix}[8]{%
	\setlength{\fboxsep}{0pt}
	\resizebox{\textwidth}{!}{%
		\fbox{\includegraphics[height=#8\linewidth]{ims/#7/#1.png}}
		\hfill
		\fbox{\includegraphics[height=#8\linewidth]{ims/#7/#2.png}}
		\hfill
		\fbox{\includegraphics[height=#8\linewidth]{ims/#7/#3.png}}
		\hfill
		\fbox{\includegraphics[height=#8\linewidth]{ims/#7/#4.png}}
		\hfill
		\fbox{\includegraphics[height=#8\linewidth]{ims/#7/#5.png}}
		\hfill
		\fbox{\includegraphics[height=#8\linewidth]{ims/#7/#6.png}}
		\vspace{2mm}
}}

%% file: 01_Introduction.tex
\section{Introduction}
\label{sec:intro}
Deep learning techniques have revolutionized the field of computer vision since their explosive appearance in the ImageNet competition~\cite{Rus+15}, where the task is to classify images into predefined categories, that is, algorithms produce one label for each input image.
Image and video segmentation, on the other hand, generate dense predictions where each pixel receives a (potentially different) output classification.
Deep learning algorithms, especially Convolutional Neural Networks (CNNs), were adapted to this scenario by removing the final fully connected layers to produce dense predictions.

Supervised techniques, those that train from manually-annotated results, are currently the best performing in many public benchmarks and challenges~\cite{Zha+17, He+17}.
In the case of image and video segmentation, the supervision is in the form of dense annotations, \ie each pixel has to be annotated in an expensive and cumbersome process.
Weakly-supervised techniques, which train from incomplete but easier-to-obtain annotations, are still significantly behind the state of the art.
Semi-automatic techniques, which need a human in the loop to produce results, are another way of circumventing the expensive training annotations but need interaction at test time, which usually comes in the form of a bounding box~\cite{DHS15} or scribbles~\cite{Lin+16} around the object of interest. How to incorporate this information at test time without introducing unacceptable lag, is also a challenge.

This paper tackles all these scenarios in a unified way and shows state-of-the-art results in all of them in a variety of benchmarks and setups. We present \longours{} (\ours{}), that obtains an object segmentation from its four extreme points~\cite{Pap+17}: the left-most, right-most, top, and bottom pixels.
Figure~1 shows an example result of our technique along with the input points provided.

In the context of semi-automatic object segmentation, we show that information from extreme clicking results in more accurate segmentations than the ones obtained from bounding-boxes (PASCAL, COCO, Grabcut) in a Grabcut-like formulation. \ours{} outperforms other methods using extreme points or object proposals (PASCAL), and provides a better input to video object segmentation (DAVIS 2016, DAVIS 2017).
\ours{} can also incorporate more points beyond the extreme ones, which further refines the quality (PASCAL).

\ours{} can also be used to obtain dense annotations to train supervised techniques.
We show that we obtain very accurate annotations with respect to the ground truth, but more importantly, that algorithms trained on the annotations obtained by our algorithm perform as good as when trained from the ground-truth ones.
If we add the cost to obtain such annotations into the equation, then training using \ours{} is significantly more efficient than training from the ground truth for a given target quality.

We perform an extensive and comprehensive set of experiments on COCO, PASCAL, Grabcut, DAVIS 2016, and DAVIS 2017, to demonstrate the effectiveness of our approach. All code, pre-trained models and pre-computed results used in this project are publicly available on \url{http://www.vision.ee.ethz.ch/~cvlsegmentation/dextr/}.

%% file: 02_Related.tex
\section{Related Work}
\label{sec:related}
\vspace{-1mm}
\paragraph{Weakly Supervised Signals for Segmentation:} Numerous alternatives to expensive pixel-level segmentation have been proposed and used in the literature. Image-level labels~\cite{Pat+15}, noisy web labels~\cite{ACP14, JOS17} and scribble-level labels~\cite{Lin+16} are some of the supervisory signal that have been used to guide segmentation methods. Closer to our approach, \cite{Bea+16} employs point-level supervision in the form of a single click to train a CNN for semantic segmentation and~\cite{Pap+17a} uses central points of an imaginary bounding box to weakly supervise object detection. Also related to our approach, \cite{DHS15, Kho+17} train semantic segmentation methods from box supervision. Recently, Papadopoulos et al. proposed a novel method for annotating objects by extreme clicks~\cite{Pap+17}. They show that extreme clicks provide additional information to a bounding box, which they use to enhance GrabCut-like object segmentation from bounding boxes. Different than these approaches, we use extreme clicking as a form of guidance for deep architectures, and show how this additional information can be used to further boost accuracy of segmentation networks, and help various applications.
\vspace{-3mm}

\paragraph{Instance Segmentation:} Several works have tackled the task of grouping pixels by object instances. Popular grouping methods provide instance segmentation in the form of automatically segmented object proposals~\cite{Hos+16, Pon+17a}. Other variants provide instance-level segmentation from a weak guiding signal in the form of a bounding box~\cite{Rot+04}. Accuracy for both groups of methods has increased by recent approaches that employ deep architectures trained on large dataset with strong supervisory signals, to learn how to produce class-agnostic masks from patches~\cite{Pin+15, Pin+16}, or from bounding boxes~\cite{Xu+17}. Our approach relates to the second group, since we utilize information from extreme clicks to group pixels of the same instance, with higher accuracy.
\vspace{-3mm}

\paragraph{Interactive Segmentation from points:} Interactive segmentation methods have been proposed in order to reduce annotation time. In this context, the user is asked to gradually refine a method by providing additional labels to the data. Grabcut~\cite{Rot+04} is one of the pioneering works for the task, segmenting from bounding boxes by gradually updating an appearance model. Our method relates with interactive segmentation using points as the supervisory signal. Click Carving~\cite{JDG16} interactively updates the result of video object segmentation by user-defined clicks. Recent methods use these ideas in the pipeline of deep architectures. iFCN~\cite{Xu+16} guides a CNN from positive and negative points acquired from the ground-truth masks. RIS-Net~\cite{Lie+17} build on iFCN to improve the result by adding local context. Our method significantly improves the results by using 4 points as the supervisory signal: the extreme points.

%% file: 03_Method.tex
\section{Method}
\label{sec:method}
\vspace{-1mm}

\begin{figure*}
\centering
\vspace{-2mm}
\includegraphics[width=.98\textwidth]{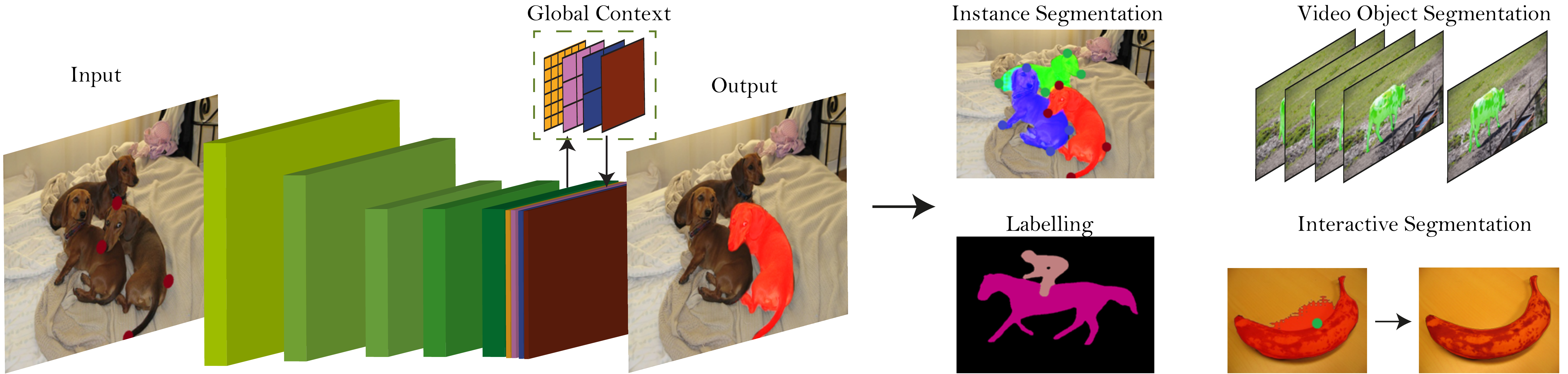}\\[1mm]
\caption{\textbf{Architecture of \ours}: Both the RGB image and the labeled extreme points are processed by the CNN to produce the segmented mask. The applicability of this method is illustrated for various tasks: Instance, Semantic, Video, and Interactive segmentation.}
\label{fig:overview}
\vspace{-4mm}
\end{figure*}
\subsection{Extreme points}
\vspace{-1mm}
One of the most common ways to perform weakly supervised segmentation is drawing a bounding box around the object of interest \cite{BoJo01,Wu+14,Rot+04,Lempitsky2009}. 
However, in order to draw the corners of a bounding box, the user has to click points outside the object, drag the box diagonally, and adjust it several times to obtain a tight, accurate bounding box. This process is cognitively demanding, with increased error rates and labelling times~\cite{Pap+17}.

Recently, Papadopoulos et al.~\cite{Pap+17} have shown a much more efficient way of obtaining a bounding box using extreme clicks, spending on average $7.2$ seconds instead of $34.5$ seconds required for drawing a bounding box around an object~\cite{Su+12}. They show that extreme clicking leads to high quality bounding boxes that are on par with the ones obtained by traditional methods. 
These extreme points belong to the top, bottom, left-most and right-most parts of the object. Extreme-clicking annotations by definition provide more information than a bounding box; they contain four points that are on the boundary of the object, from which one can easily obtain the bounding-box. We use extreme points for object segmentation leveraging their two main outcomes: the points and their inferred bounding box.

\subsection{Segmentation from Extreme Points}
The overview of our method is shown in Figure~\ref{fig:overview}. The annotated extreme points are given as a guiding signal to the input of the network. To this end, we create a heatmap with activations in the regions of extreme points. We center a 2D Gaussian around each of the points, in order to create a single heatmap.
The heatmap is concatenated with the RGB channels of the input image, to form a 4-channel input for the CNN. In order to focus on the object of interest, the input is cropped by the bounding box, formed from the extreme point annotations. To include context on the resulting crop, we relax the tight bounding box by several pixels.
After the pre-processing step that comes exclusively from the extreme clicks, the input consists of an RGB crop including an object, plus its extreme points.

We choose \textit{ResNet-101}~\cite{He+16} as the backbone of our architecture, as it has been proven successful in a variety of segmentation methods~\cite{Che+17, He+17}. 
We remove the fully connected layers as well as the max pooling layers in the last two stages to preserve acceptable output resolution for dense prediction, and we introduce atrous convolutions in the last two stages to maintain the same receptive field. After the last \textit{ResNet-101} stage, we introduce a pyramid scene parsing module~\cite{Zha+17} to aggregate global context to the final feature map. Initializing the weights of the network from pre-training on ImageNet has been proven beneficial for various tasks~\cite{LSD15, XiTu17, He+17}. For most experiments, we use the provided Deeplab-v2 model pre-trained on ImageNet, and fine-tuned on PASCAL for semantic segmentation.
 
The output of the CNN is a probability map representing whether a pixel belongs to the object that we want to segment or not.
The CNN is trained to minimize the standard cross entropy loss, which takes into account that different classes occur with different frequency in a dataset:
\vspace{-2mm}
\begin{equation}
\mathcal{L} = \sum_{j \in Y}{w_{y_j} C\left(y_j, \hat{y}_j\right)}, \qquad j \in {1,...,|Y|}
\vspace{-2mm}
\end{equation}
where $w_{y_j}$ depends on the label $y_j$ of pixel $j$. In our case we define $w_{y_j}$ with $ y_j \in \left\lbrace 0, 1\right\rbrace$ as the inverse normalized frequency of labels inside the minibatch. $C(.)$ indicates the standard cross-entropy loss between the label and the prediction $\hat{y}_j$. The balanced loss has proven to perform very well in boundary detection~\cite{XiTu17, Man+17}, where the majority of the samples belong to the background class. We note that our method is trained from strong mask-level supervision, on publicly available datasets, using the extreme points as a guiding signal to the network.

In order to segment an object, our method uses a object-centered crop, therefore there is a much higher number of samples belonging to the foreground than to the background and the use of a balanced loss proves to be beneficial. 

Alternatives for each of the components used in our final model have been studied in an ablation analysis, and a detailed comparison can be found in Section~\ref{sec:exp:ablation}.

\subsection{Use cases for \ours}
\vspace{-1mm}
\paragraph{Class-agnostic Instance Segmentation:}
One application of \ours{} is class-agnostic instance segmentation. In this task, we click on the extreme points of an object in an image, and we obtain a mask prediction for it. The selected object can be of any class, as our method is class agnostic.  

In Section~\ref{sec:exp:instance}, we compare our method with the state of the art in two different datasets, PASCAL and Grabcut, where we improve current results. We also analyse the generalization of our method to other datasets and to unseen categories. We conclude positive results in both experiments: the performance drop for testing on a different dataset than the one used for training is very small and the result achieved is the same whether the class has been seen during training or not.\vspace{-3mm}

\paragraph{Annotation:}
The common annotation pipeline for segmentation can also be assisted by \ours{}. In this framework, instead of detailed polygon labels, the workload of the annotator is reduced to only providing the extreme points of an object, and \ours{} produces the desired segmentation. In this pipeline, the labelling cost is reduced by a factor of 10 (from 79 seconds needed for a mask, to 7.2 seconds needed for the extreme clicks)~\cite{Pap+17}.

In Section~\ref{sec:exp:annotation}, the quality of the produced masks are validated when used to train a semantic segmentation algorithm. We show that our method produces very accurate masks and the results trained on them are on par with those trained on the ground-truth annotations in terms of quality, with much less annotation budget.\vspace{-3mm}

\paragraph{Video Object Segmentation:}
\ours{} can also improve the pipeline of video object segmentation. We focus on the semi-supervised setting where methods use one or more masks as inputs to produce the segmentation of the whole video. Our aim is to replace the costly per pixel annotation masks by the masks produced by our algorithm after the user has selected the extreme points of a certain object, and re-train strongly supervised state-of-the-art video segmentation architectures.

In Section~\ref{sec:exp:vos}, we provide results on two different dataset: DAVIS-2016 and DAVIS-2017. We conclude that state-of-the-art results can be achieved reducing the annotation time by a factor of 5. Moreover, for almost any specific annotation budget, better results can be obtained using a higher number of masks produced by our algorithm rather than expensive per-pixel annotated masks.  \vspace{-3mm}

\paragraph{Interactive Object Segmentation:}
The pipeline of \ours{} can also be used in the frame of interactive segmentation from points~\cite{Xu+16,Xu+17}. We work on the case where the user labels the extreme points of an object, but is nevertheless not satisfied with the obtained results. The natural thing to do in such case is to annotate an extra point (not extreme) in the region that segmentation fails, and expect for a refined result. Given the nature of extreme points, we expect that the extra point also lies in the boundary of the object. 

To simulate such behaviour, we first train \ours{} on a first split of a training set of images, using the 4 extreme points as input. For the extra point, we infer on an image of the second split of the training set, and compute the accuracy of its segmentation. If the segmentation is accurate (eg. $IoU \ge 0.8$), the image is excluded from further processing. In the opposite case ($IoU < 0.8$), we select a fifth point in the erroneous area. To simulate human behaviour, we perturbate its location and we train the network with 5 points as input. Results presented in Section~\ref{sec:exp:interactive} indicate that it is possible to recover performance on the difficult examples, by using such interactive user input.

%% file: 04a_Ablation.tex
\section{Experimental Validation}
\label{sec:exp}
Our method is extensively validated on five publicly available databases: PASCAL~\cite{Eve+12}, COCO~\cite{Lin+14}, DAVIS 2016~\cite{Per+16}, DAVIS 2017~\cite{Pon+17}, and Grabcut~\cite{Rot+04}, for various experimental setups that show its applicability and generalization capabilities.
We use \ours{} trained on PASCAL (augmented by the labels of SBD~\cite{Har+11} following the common practice - 10582 images), unless indicated differently.
Some implementation details are given in Section~\ref{sec:impl_det}.
We then perform an ablation study to separately validate all components of our method in Section~\ref{sec:exp:ablation}.
Class-agnostic instance segmentation experiments from extreme points are presented in Section~\ref{sec:exp:instance}, whereas Sections~\ref{sec:exp:annotation} and~\ref{sec:exp:vos} are dedicated to how \ours{} contributes to segmentation annotation and video object segmentation pipelines, respectively.
Section~\ref{sec:exp:interactive} presents our method as an interactive segmenter from points.

\subsection{Implementation Details}
\label{sec:impl_det}
\paragraph{Simulated Extreme Points:}
In~\cite{Pap+17}, extreme points in PASCAL were obtained by crowd-sourcing.
We used their collected extreme points when experimenting on the same dataset, and collected new extreme points by humans in DAVIS 2016.
To experiment on COCO, on which it was not feasible to collect extreme points by human annotators, we simulate them by taking the extreme points of the ground-truth masks jittered randomly by up to 10 pixels.
\vspace{-1mm}

\paragraph{Training and testing details:}
\ours{} is trained on PASCAL 2012 Segmentation for 100 epochs or on COCO 2014 training set for 10 epochs.
The learning rate is set to $10^{-8}$, with momentum of $0.9$ and weight decay of $5*10^{-4}$.
A mini-batch of 5 objects is used for PASCAL, whereas for COCO, due to the large size of the database, we train on 4 GPUs with an effective batch size of 20.
Training on PASCAL takes approximately 20 hours on a Nvidia Titan-X GPU, and 5 days on COCO.
Testing the network is fast, requiring only 80 milliseconds.

\subsection{Ablation Study}
\label{sec:exp:ablation}
The following sections show a number of ablation experiments in the context of class-agnostic instance segmentation to quantify the importance of each of the components of our algorithm and to justify various design choices. Table~\ref{tab:ablation_vs} summarizes these results. We use PASCAL VOC 2012 val set for the evaluation.\vspace{-1mm}

\paragraph{Architecture:}
We use \textit{ResNet-101} as the backbone architecture, and compare two different alternatives. The first one is a straightforward fully convolutional architecture (Deeplab-v2~\cite{Che+17}) where the fully connected and the last two max pooling layers are removed, and the last two stages are substituted with dilated (or atrous) convolutions. This keeps the size of the prediction in reasonable limits ($8\times$ lower than the input).
We also tested a region-based architecture, similar to Mask R-CNN~\cite{He+17}, with a re-implementation of the ResNet-101-C4 variant~\cite{He+17}, which uses the fifth stage (C5) for regressing a mask from the Region of Interest (RoI), together with the re-implementation of the RoI-Align layer. For more details please refer to~\cite{He+17}. In the first architecture, the input is a patch around the object of interest, whereas in the latter the input is the full image, and cropping is applied at the RoI-Align stage. Deeplab-v2 performs +3.9\% better. We conclude that the output resolution of ResNet-101-C4 ($28\!\times\!28$) is inadequate for the level of detail that we target.\vspace{-1mm}

\paragraph{Bounding boxes vs.\ extreme points:}
We study the performance of Deeplab-v2 as a foreground-background classifier given a bounding box compared to the extreme points.
In the first case, the input of the network is the cropped image around the bounding box plus
a margin of 50 pixels to include some context.
In the second case, the extreme points are fed together in a fourth channel of the input to guide the segmentation.
Including extreme points to the input increases performance by +3.1\%, which suggest that they are a source of very valuable information that the network uses additionally to guide its output.\vspace{-1mm}

\paragraph{Loss:}
For the task of class-agnostic instance segmentation, we compare two binary losses, i.e. the standard cross-entropy and a class-balanced version of it, where the loss for each class in the batch is weighted by its inverse frequency. Class-balancing the loss gives more importance to the less frequent classes, and has been successful in various tasks~\cite{XiTu17, Cae+17}.
\ours{} also performs better when the loss is balanced, leading to a performance boost of +3.3\%.\vspace{-1mm}

\paragraph{Full image vs.\ crops:}
Having the extreme points annotated allows for focusing on specific regions in an image, cropped by the limits specified by them. In this experiment, we compare how beneficial it is to focus on the region of interest, rather than processing the entire image. To this end, we crop the region surrounded by the extreme points, relaxing it by 50 pixel for increased context and compare it against the full image case. We notice that cropping increases performance by +7.9\%, and is especially beneficial for the small objects of the database.
This could be explained by the fact that cropping eliminates the scale variation on the input.
Similar findings have been reported for video object segmentation by~\cite{DAVIS2017-1st}.\vspace{-1mm}

\paragraph{Atrous spatial pyramid (ASPP) vs.\ pyramid scene parsing (PSP) module:}
Pyramid Scene Parsing Network~\cite{Zha+17} steps on the Deeplab-v2~\cite{Che+17} architecture to further improve results on semantic segmentation. Their main contribution was a global context module (PSP) that employs global features together with the local features for dense prediction. We compare the two network heads, the original ASPP~\cite{Che+17}, and the recent PSP module~\cite{Zha+17} for our task. The increased results of the PSP module (+2.3\%) indicate that the PSP module builds a global context that is also useful in our case.\vspace{-1mm}

\paragraph{Manual vs.\ simulated extreme points:} 
In this section we analyze the differences between the results obtained by \ours{} when we input either human-provided extreme points or our simulated ones, to check that the conclusions we draw from the simulations will still be valid in a realistic use case with human annotators.
We do so in the two datasets where we have \textit{real} extreme points from humans. The first one is a certain subset of PASCAL 2012 Segmentation and SBD (5623 objects) with extreme points from~\cite{Pap+17}, which we refer to as PASCAL$_{EXT}$ and DAVIS 2016, for which we crowdsourced the extreme point annotations. The annotation time for the latter (average of all 1376 frames of the validation set) was 7.5 seconds per frame,
in line with~\cite{Pap+17} (7.2~s.\ per image).
Table~\ref{tab:manual_vs_simulated_extremepoints} shows that the results are indeed comparable when using both type of inputs. 
The remainder of the paper uses the simulated extreme points except when otherwise specified.

\begin{table}[t]
\centering
\rowcolors{2}{white}{rowblue}
\resizebox{\linewidth}{!}{%
\begin{tabular}{lcc}
\toprule
Method       &        PASCAL$_{EXT}$      &     DAVIS 2016 \\
\midrule
Manual extreme points	&   80.1 & \ 80.9\\
Simulated extreme points$\qquad\quad$			&  85.1  & \ 79.5\\
\bottomrule
\end{tabular}}
\vspace{2mm}
\caption{\textbf{Manual vs. simulated extreme points}: Intersection over Union (IoU) of the \ours{} results when using manual or simulated extreme points as input.}
\label{tab:manual_vs_simulated_extremepoints}
\end{table}
\vspace{-2mm}

\paragraph{Distance-map vs.\ fixed points:}
Recent works~\cite{Xu+16, Xu+17, Lie+17} that focus on segmentation from (not-extreme) points use the distance transform of positive and negative annotations as an input to the network, in order to guide the segmentation. We compare with their approach by substituting the fixed Gaussians to the distance transform of the extreme points. We notice a performance drop of -1.3\%, suggesting that using fixed Gaussians centered on the points is a better representation when coupled with extreme points.
In Section~\ref{sec:exp:instance} we compare to such approaches, showing that extreme points provide a much richer guidance than arbitrary points on the foreground and the background of an object.

\paragraph{Summary}
\label{sec:exp:ablation:summary}
% Table~\ref{tab:ablation_vs} summarizes the main ablated results that have been discussed above, analyzing all components. 

\begin{table}[h]
\centering
\rowcolors{2}{white}{rowblue}
\resizebox{0.82\linewidth}{!}{%
\begin{tabular}{ccr}
\toprule
Component \#1       & $\quad$ Component \#2		$\quad$    & Gain in IoU\\
\midrule
Region-based		& \bf\ Deeplab-v2 		& $+3.9\%\quad$     \\ 
Bounding Boxes 		& \bf\ Extreme Points 	& $+3.1\%\quad$	 \\
Cross Entropy		& \bf\ Balanced BCE 	& $+3.3\%\quad$	 \\
Full Image			& \bf\ Crop on Object	& $+7.9\%\quad$     \\
ASPP				& \bf\ PSP 				& $+2.3\%\quad$	 \\
\bf Fixed Points	& 	 \ Distance Map		& $-1.3\%\quad$	 \\
\bottomrule

\end{tabular}}
\vspace{2mm}
\caption{\textbf{Ablation study}: Comparative evaluation between different choices in various components of our system.
Mean IoU over all objets in PASCAL VOC 2012 val set.}
\label{tab:ablation_vs}
\end{table}

\begin{table}[b]
\centering
\rowcolors{2}{white}{rowblue}
\resizebox{\linewidth}{!}{%
\begin{tabular}{lcr}
\toprule
Variant       & IoU (\%) & Gain\\
\midrule
Full Image (Deeplab-v2 + PSP + Extreme Points)		&    82.6 &     	   \\ 
\midrule
Crop on Object (Deeplab-v2)							&    85.1 & +2.5\%     \\ 
+ PSP												&    87.4 & +2.3\%     \\ 
+ Extreme Points 									&    90.5 & +3.1\%     \\ 
\bf+ SBD data (Ours) $\qquad\qquad\qquad$			& \bf91.5 & +1.0\%     \\ 
\bottomrule
\end{tabular}}
\vspace{2mm}
\caption{\textbf{Ablation study}: Building performance in PASCAL VOC 2012 val set.}
\label{tab:ablation}
\end{table}

Table~\ref{tab:ablation} illustrates the building blocks that lead to the best performing variant for our method. All in all, we start by a Deeplab-v2 base model working on bounding boxes. We add the PSP module (+2.3\%), the extreme points in the input of the network (+3.1\%), and more annotated data from SBD (+1\%) to reach maximum accuracy. The improvement comes mostly because of the guidance from extreme points, which highlights their importance for the task.

%% file: 04b_InstanceSeg.tex
\begin{figure*}
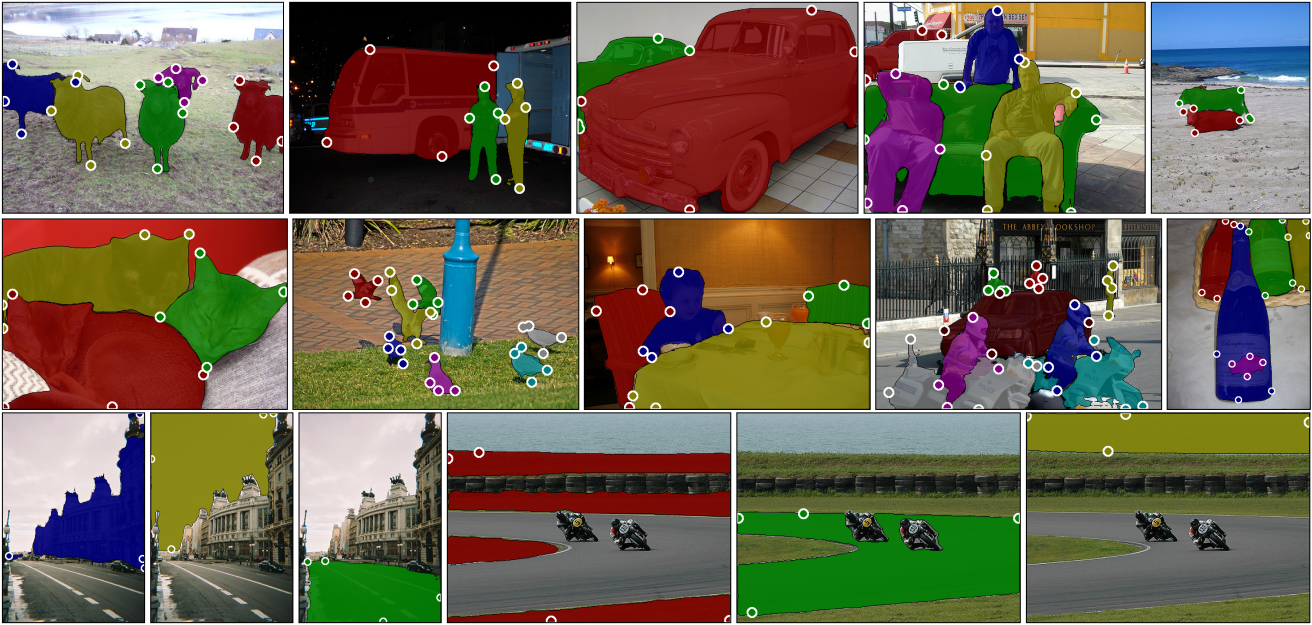

\centering
\showonerow{2009_003666}{2008_006554}{2008_003577}{2010_001767}{2007_000464}{instance}{0.2} \\[0.3mm]
\showonerow{2010_000683}{2009_004568}{2009_003071}{2009_000825}{2009_000121}{instance}{0.2}
\showonerowsix{2008_000253-building}{2008_000253-sky}{2008_000253-road}{2008_001727-grass}{2008_001727-road}{2008_001727-water}{stuff}{0.2}
\caption{\textbf{Qualitative results by \ours{} on PASCAL:} Each instance with the simulated extreme points used as input and the resulting mask overlayed. The bottom row shows results on PASCAL Context stuff categories.}
\label{fig:qualitative}
\vspace{-3mm}
\end{figure*}

\subsection{Class-agnostic Instance Segmentation}
\label{sec:exp:instance}

\paragraph*{Comparison to the State of the Art in PASCAL:} 
We compare our method against state-of-the-art class-agnostic instance segmentation methods in Table~\ref{tab:instance_comparison}.
\ours{} gets a boost of +6.5\% with respect to using the \textit{grabcut-based} method of~\cite{Pap+17} from extreme points.

\begin{table}[h]
\centering
\rowcolors{2}{white}{rowblue}
\resizebox{0.85\linewidth}{!}{%
\begin{tabular}{lc}
\toprule
Method     &  IoU\\
\midrule
Sharpmask~\cite{Pin+16} from bounding box  $\qquad\qquad$& 69.3\%   \\
Sharpmask~\cite{Pin+16} upper bound    & 78.0\%   \\
\cite{Pap+17} from extreme points &73.6\%\\
{\bf Ours}  from extreme points   & \bf 80.1\%  \\
\bottomrule
\end{tabular}}
\vspace{2mm}
\caption{\textbf{Comparison on PASCAL$_{EXT}$}: IoU of our results against class-agnostic instance segmentation methods, on the objects annotated by~\cite{Pap+17} to be able to compare to them.}
\label{tab:instance_comparison}
\vspace{-1mm}
\end{table}

We then compare to two other \textit{baselines} using SharpMask~\cite{Pin+16},
the state-of-the-art object proposal technique.
In the first row, we evaluate the proposal (out of 1000) whose bounding box best overlaps with the ground-truth bounding box, mimicking a naive algorithm to segment boxes from proposals.
The second row shows the upper bound of SharpMask, that is, the best proposal against the ground truth, selected by an oracle.
Both approaches are well below our result (-10.8\% and -2.1\%).
Figure~\ref{fig:qualitative} illustrates some results obtained by our method on PASCAL.
\vspace{-2mm}

\paragraph*{Comparison to the State of the Art on the Grabcut dataset:} 
We use our best PASCAL model and we test it in the Grabcut dataset~\cite{Rot+04}.
This dataset contains 50 images, each with one annotated object from various categories, some of them not belonging to any of the PASCAL ones (banana, scissors, kangaroo, etc.).
The evaluation metric is the error rate: the percentage of misclassified pixels within the bounding boxes provided by~\cite{Lempitsky2009}.
Table~\ref{tab:grabut} shows the results, where \ours{} achieves 2.3\% error rate, 1.1\% below the runner up (or a 32\% relative improvement).
\vspace{-1mm}

\begin{table}[h]
\centering
\rowcolors{2}{white}{rowblue}
\resizebox{.75\linewidth}{!}{%
\begin{tabular}{lc}
\toprule
Method       & Error Rate (\%)\\
\midrule
GrabCut~\cite{Rot+04}			&    \ 8.1   \\
KernelCut~\cite{Tang2015}		&    \ 7.1   \\
OneCut~\cite{Tang2013}			&    \ 6.7   \\
\cite{Pap+17} from extreme points$\qquad\qquad$	&	 \ 5.5   \\
BoxPrior~\cite{Lempitsky2009}			&    \ 3.7   \\
MILCut~\cite{Wu+14}			&    \ 3.6   \\
DeepGC~\cite{Xu+17}			&    \ 3.4   \\
{\bf Ours} from extreme points			& \bf\ 2.3   \\
\bottomrule
\end{tabular}}
\vspace{2mm}
\caption{\textbf{Comparison on the Grabcut dataset}: Error rates compared to the state-of-the-art techniques.}
\label{tab:grabut}
\vspace{-1mm}
\end{table}

\paragraph*{Generalization to unseen categories and across datasets:}
Table~\ref{tab:unseen_classes} shows our results when trained on a certain dataset (first column), and tested in another one or certain categories (second column).
In order to make a fair comparison, all the models are pre-trained only on ImageΝet~\cite{Rus+15} for image labeling and trained on the specified dataset for category-agnostic instance segmentation.
The first two rows show that our technique is indeed class agnostic, since the model trained on PASCAL achieves roughly the same performance in COCO mini-val (MVal) regardless of the categories tested.
The remaining rows shows that \ours{} also generalizes very well across datasets, since differences are around only \mbox{2\%} of performance drop.

\begin{table}[h]
\centering
\rowcolors{2}{white}{rowblue}
\resizebox{\linewidth}{!}{%
\begin{tabular}{clcc}
\toprule
&Train       &  Test       & IoU\\
\midrule

\cellcolor{white}&PASCAL 	&  COCO MVal w/o PASCAL classes &   80.3\%   \\
\multirow{-2}{*}{\thead{\cellcolor{white}Unseen\\\cellcolor{white}categories}}&PASCAL  &  COCO MVal only PASCAL classes    &79.9\%   \\
\midrule
\cellcolor{white}&PASCAL  &  COCO MVal    &80.1\%   \\
&COCO   &  COCO MVal    &82.1\%   \\
\cmidrule{2-4}
&\cellcolor{rowblue}COCO   &  \cellcolor{rowblue}PASCAL    & \cellcolor{rowblue}87.8\%   \\
\multirow{-4}{*}{\thead{\cellcolor{white}Dataset\\\cellcolor{white}generalization}}\cellcolor{white}&\cellcolor{white}PASCAL 	&  \cellcolor{white}PASCAL & \cellcolor{white}89.8\%   \\
\bottomrule

\end{tabular}}
\vspace{2mm}
\caption{\textbf{Generalization to unseen classes and across datasets}: Intersection over union results of training in one setup and testing on another one. MVal stands for mini-val.}
\label{tab:unseen_classes}
\vspace{-3mm}
\end{table}

\paragraph*{Generalization to background (stuff) categories:}
In order to verify the performance of DEXTR in ``background'' classes, we trained a model using the background labels of PASCAL Context~\cite{Mot+14} (road, sky, sidewalk, building, wall, fence, grass, ground, water, floor, ceiling, mountain, and tree). Qualitative results (Figure~\ref{fig:qualitative} last row) suggest that our method generalizes to background classes as well. Quantitatively, we achieve a mIoU of 81.75\% in PASCAL-Context validation set, for the aforementioned classes.

%% file: 04c_Annotation.tex
\subsection{Annotation}
\label{sec:exp:annotation}
As seen in the previous section, \ours{} is able to generate high-quality class-agnostic masks given only extreme points as input.
The resulting masks can in turn be used to train other deep architectures for other tasks or datasets, that is, we use extreme points as a way to annotate a new dataset with object segmentations.
In this experiment we compare the results of a semantic segmentation algorithm trained on either the ground-truth masks or those generated by \ours{} (we combine all per-instance segmentations into a per-pixel semantic classification result).

Specifically, we train \ours{} on COCO and use it to generate the object masks of PASCAL train set, on which we train
Deeplab-v2~\cite{Che+17}, and the PSP~\cite{Zha+17} head as the semantic segmentation network.
To keep training time manageable, we do not use multi-scale training/testing.
We evaluate the results on the PASCAL 2012 Segmentation val set, and measure performance by the standard mIoU measure
(IoU per-category and averaged over categories).

\begin{figure}[h]
\centering
\resizebox{\linewidth}{!}{\begin{tikzpicture}[/pgfplots/width=0.85\linewidth, /pgfplots/height=0.64\linewidth, /pgfplots/legend pos=south east]
    \begin{axis}[ymin=20,ymax=85,xmin=0.12,xmax=4000,enlargelimits=false,
        xlabel=Annotation budget (hours),
        ylabel= mIoU (\%),
		font=\scriptsize,
        grid=both,
		grid style=dotted,
        xlabel shift={-2pt},
        ylabel shift={-5pt},
        xmode=log,
        legend columns=1,
        %transpose legend,
        legend style={/tikz/every even column/.append style={column sep=3mm}},
        minor ytick={10,25,...,90},
        ytick={10,20,...,90},
		yticklabels={10,20,...,90},
	    xticklabels={0.1,1,10,100, 1000},
        legend pos= south east
        ]

        \addplot+[smooth,red,mark=none, line width=1.5] table[x=hours,y=mIoU] {data/semseg/Extreme_by_budget.txt};
        \addlegendentry{Ours}
        \label{fig:qual_vs_time_budget:ours}

        \addplot+[smooth,blue,mark=none, line width=1] table[x=hours,y=mIoU] {data/semseg/GT_by_budget.txt};
        \addlegendentry{GT}
        \label{fig:qual_vs_time_budget:gt}
        
        	\addplot+[cyan,fill=cyan,mark=*, mark size=1.5,only marks, mark options={fill=cyan}] coordinates{(2937, 81.3454)};
        \addlegendentry{PSP~\cite{Zha+17}}
        \label{fig:qual_vs_time_budget:psp}

    \end{axis}
\end{tikzpicture}

\begin{tikzpicture}[/pgfplots/width=0.85\linewidth, /pgfplots/height=0.64\linewidth, /pgfplots/legend pos=south east]
    \begin{axis}[ymin=20,ymax=85,xmin=10,xmax=150000,enlargelimits=false,
        xlabel=Number of annotated images,
        ylabel= mIoU (\%),
		font=\scriptsize,
        grid=both,
		grid style=dotted,
        xlabel shift={-2pt},
        ylabel shift={-5pt},
        xmode=log,
        legend columns=1,
        %transpose legend,
        legend style={/tikz/every even column/.append style={column sep=3mm}},
        minor ytick={10,25,...,90},
        ytick={10,20,...,90},
		yticklabels={10,20,...,90},
	    xticklabels={10,100,1000,10000,100000},
        legend pos= south east
        ]

        \addplot+[smooth,red,mark=none, line width=1.5] table[x=num_images,y=mIoU] {data/semseg/Extreme_by_nImgs.txt};
        \addlegendentry{Ours}
        \label{fig:qual_vs_time_nImgs:ours}

        \addplot+[smooth,blue,mark=none, line width=1] table[x=num_images,y=mIoU] {data/semseg/GT_by_nImgs.txt};
        \addlegendentry{GT}
        \label{fig:qual_vs_time_nImgs:gt}

        % 123286 COCO + 10582 PASCAL = 133868 images
        	\addplot+[cyan,fill=cyan,mark=*, mark size=1.5,only marks, mark options={fill=cyan}] coordinates{(133868, 81.3454)};
        \addlegendentry{PSP~\cite{Zha+17}}
        \label{fig:qual_vs_time_nImgs:psp}
           
    \end{axis}
\end{tikzpicture}}
\vspace{-5mm}
   \caption{\textbf{Quality vs.\ annotation budget}: mIoU for semantic segmentation on PASCAL val set trained on our masks or the input, as a function of annotation budget (left) and the number of annotated images (right).}
\label{fig:qual_vs_time_budget}
   \vspace{-3mm}
\end{figure}
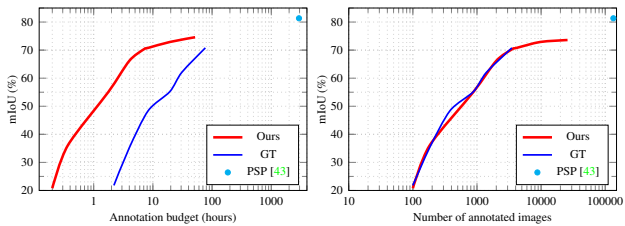

Figure~\ref{fig:qual_vs_time_budget} shows the results with respect to the annotation budget (left) and the number of images (right).
For completeness, we also report the results of PSPNet~\cite{Zha+17} (\ref{fig:qual_vs_time_budget:psp}) by evaluating the model provided by the authors (pre-trained on COCO, with multi-scale training and testing).
The results trained on \ours{}'s masks are significantly better than those trained from the ground truth on the same budget
(e.g.\ 70\% IoU at 7-minute annotation time vs.\ 46\% with the same budget, or 1h10 instead of 7 minutes to reach the same 70\% accuracy).
\ours{}'s annotations reach practically the same performance than ground truth when given the same number of annotated images.

%% file: 04d_VideoSeg.tex
\subsection{Video Object Segmentation}
\label{sec:exp:vos}
We test \ours{} also for Video Object Segmentation on the DAVIS datasets~\cite{Per+16,Pon+17}.
We focus on the semi-supervised setting \ie the mask in one or more frames of the object that we want to segment is given as input to the algorithm, and as before we will compare the results obtained from the masks obtained by \ours{} or the ground truth having a certain annotation budget.
We assume that the annotation time of the DAVIS masks is the same than that of COCO~\cite{Lin+14} ($79$ seconds per instance), despite the former are significantly more accurate.

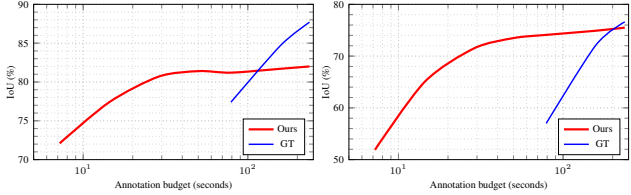
\begin{figure}[h]
\centering
\resizebox{\linewidth}{!}{\begin{tikzpicture}[/pgfplots/width=1\linewidth, /pgfplots/height=0.64\linewidth, /pgfplots/legend pos=south east]
    \begin{axis}[ymin=70,ymax=90,xmin=5,xmax=250,enlargelimits=false,
        xlabel=Annotation budget (seconds),
        ylabel= IoU (\%),
		font=\scriptsize,
        grid=both,
		grid style=dotted,
        xlabel shift={-2pt},
        ylabel shift={-5pt},
        xmode=log,
        legend columns=1,
        %transpose legend,
        legend style={/tikz/every even column/.append style={column sep=3mm}},
        minor ytick={70,71,...,90},
        ytick={70,75,...,90},
		yticklabels={70,75,...,90},
        legend pos = south east
        ]

        \addplot+[smooth,red,mark=none, line width=1.5] table[x=seconds,y=mIoU] {data/videoseg/davis_2016_extreme.txt};
        \addlegendentry{Ours}
        \label{fig:davis_2016:ours}

        \addplot+[smooth,blue,mark=none, line width=1] table[x=seconds,y=mIoU] {data/videoseg/davis_2016_gt.txt};
        \addlegendentry{GT}
        \label{fig:davis_2016:gt}

    \end{axis}
\end{tikzpicture}

\begin{tikzpicture}[/pgfplots/width=1\linewidth, /pgfplots/height=0.64\linewidth, /pgfplots/legend pos=south east]
    \begin{axis}[ymin=50,ymax=80,xmin=5,xmax=250,enlargelimits=false,
        xlabel=Annotation budget (seconds),
        ylabel= IoU (\%),
		font=\scriptsize,
        grid=both,
		grid style=dotted,
        xlabel shift={-2pt},
        ylabel shift={-5pt},
        legend columns=1,
        xmode=log,
        %transpose legend,
        legend style={/tikz/every even column/.append style={column sep=3mm}},
        minor ytick={50,52,...,80},
        ytick={50,60,...,80},
		yticklabels={50,60,...,80},
        legend pos= south east
        ]

        \addplot+[smooth,red,mark=none, line width=1.5] table[x=seconds,y=mIoU] {data/videoseg/davis_2017_extreme.txt};
        \addlegendentry{Ours}
        \label{fig:davis_2017:ours}

        \addplot+[smooth,blue,mark=none, line width=1] table[x=seconds,y=mIoU] {data/videoseg/davis_2017_gt.txt};
        \addlegendentry{GT}
        \label{fig:davis_2017:gt}
    \end{axis}
\end{tikzpicture}}
\vspace{-5mm}
   \caption{\textbf{Quality vs.\ annotation budget in video object segmentation}: OSVOS' performance when trained from the masks of \ours{} or the ground truth, on DAVIS 2016 (left) and on DAVIS 2017 (right).}
   \label{fig:davis}
   \vspace{-3mm}
\end{figure}

We use OSVOS~\cite{Cae+17}, as a state-of-the-art semi-supervised video object segmentation technique, which heavily relies on the appearance of the annotated frame, and their code is publicly available. 
Figure~\ref{fig:davis} (left) shows the performance of OSVOS in DAVIS 2016~\cite{Per+16} trained on the ground truth mask~(\ref{fig:davis_2016:gt}) or the masks generated by \ours{} from extreme points~(\ref{fig:davis_2016:ours}).
We reach the same performance as using one ground-truth annotated mask with an annotation budget 5 times smaller.
Once we train with more than one ground-truth annotated mask, however, even though we can generate roughly ten times more masks, we cannot achieve the same accuracy.
We believe this is so because DAVIS 2016 sequences have more than one semantic instance per mask while we only annotate a global set of extreme points, which confuses \ours{}.

To corroborate this intuition, we perform the same experiment in DAVIS 2017~\cite{Pon+17}, where almost every mask contains only one instance. Figure~\ref{fig:davis} (right) shows that the performance gap with respect to using the full ground-truth mask is much smaller than in DAVIS 2016.
Overall, we conclude that \ours{} is also very efficient to reduce annotation time in video object segmentation.

%% file: 04e_Interactive.tex
\subsection{Interactive Object Segmentation}
\label{sec:exp:interactive}
\paragraph{\ours{} for Interactive Segmentation:}
We experiment on PASCAL VOC 2012 segmentation for interactive object segmentation.
We split the training dataset into two equal splits.
Initially, we train \ours{} on the first split and test on the second.
We then focus on the objects with inaccurate segmentations, i.e.\ IoU$<$0.8, to simulate the ones on which a human - unsatisfied with the result - would mark a fifth point. The extra point would lie on the boundary of the erroneous area (false positive or false negative), which we simulate as the boundary point closest to the highest error. 
From the perspective of network training, this can be interpreted as
Online Hard Example Mining (OHEM)~\cite{SGG16}, where one only needs to back-propagate gradients for the training examples that lead to the highest losses.
Results are presented in Table~\ref{tab:interactive}.

\begin{table}[h]
\centering
\rowcolors{2}{white}{rowblue}
\resizebox{.9\linewidth}{!}{%
\begin{tabular}{lcccc}
\toprule
Trained on	& 4 points       & 4 points-all 	& 5 points	& 5 points + OHEM \\
\midrule
IoU		&59.6\%		 & 69.0\%		&	69.2\%	& 	\bf73.2\%   \\
\bottomrule
\end{tabular}}
\vspace{2mm}
\caption{\textbf{Interactive Object Segmentation Evaluation}: Average IoU on difficult cases of PASCAL VOC 2012 validation dataset.}
\label{tab:interactive}
\end{table}

We first select the objects that lead to poor performance (IoU$<$0.8) when applying the network trained on the first split. We report the average IoU on them (338 objects - 59.6\%).
Using the network trained further on the hard examples, with a fifth boundary point, performance increases to 73.2\% (``5 points + OHEM'').
 
Since the increased performance is partially due to the increased amount of training data (first split + hard examples of the second split), we need to disentangle the two sources of performance gain. To this end, we train \ours{} on 4 points, by appending the hard examples of the second split to the first split of our training set (``4 points-all'').

Results suggest that \ours{} learns to handle more input information given interactively in the form of boundary clicks, to improve results of poorly segmented difficult examples (+4.2\%).
Interestingly, OHEM is a crucial component for improving performance: without it the network does not focus on the difficult examples (only 11\% of objects of the second training split are hard examples), and fails to improve on the erroneous region indicated by the fifth boundary point (``5 points'').
\vspace{-2mm}

\paragraph{Comparison to the State of the Art:}
We compare against the state-of-the-art in interactive segmentation by considering extreme points as 4 clicks. 
Table~\ref{tab:interactive_pascal} shows the number of clicks that each method needs to reach a certain performance, as well as their performance when the input is 4 clicks, in PASCAL and the Grabcut dataset. \ours{} reaches about 10\% higher performance at 4 clicks than the best competing method, and reaches 85\% or 90\% quality with fewer clicks. 
This further demonstrates the enhanced performance of the CNN, when guided by extreme points.

\begin{table}[h]
\centering
\rowcolors{3}{rowblue}{white}
\resizebox{1\linewidth}{!}{%
\begin{tabular}{l@{\hspace{5mm}}cc@{\hspace{5mm}}cc}
\toprule
       & \multicolumn{2}{c}{Number of Clicks}	&	\multicolumn{2}{c}{IoU (\%) @ 4 clicks} \\
Method 			& PASCAL@85\%		& Grabcut@90\%	& PASCAL		& Grabcut\\
\midrule
GraphCut~\cite{BoJo01}			&    \ $>20$   	&	\ $>20$	&    \	41.1	&	\ 59.3		\\
Geodesic matting~\cite{BaSa09}	&	 \ $>20$		&	\ $>20$	&    \  45.9	&	\ 55.6	\\
Random walker~\cite{Gra06}		&    \ 16.1		&	\ 15		&    \	55.1	&	\ 56	.9	\\
iFCN	~\cite{Xu+16}				&    \ 8.7		&	\ 7.5	&    \	75.2	&	\ 84.0	\\
RIS-Net~\cite{Lie+17}			&    \ 5.7   	&	\ 6.0	&    \	80.7&	\ 85.0		\\
\bf Ours							& \bf\ 4.0		&\bf\ 4.0	& \bf\	91.5&\bf\ 94.4	\\
\bottomrule

\end{tabular}}
\vspace{2mm}
\caption{\textbf{PASCAL and Grabcut Dataset evaluation:} Comparison to interactive segmentation methods in terms of number of clicks to reach a certain quality and in terms of quality at 4 clicks.}
\label{tab:interactive_pascal}
\vspace{-2mm}
\end{table} 

To the meticulous reader, please note that the difference in performance of \ours{} between Table~\ref{tab:interactive_pascal} (91.5\%) and Table~\ref{tab:manual_vs_simulated_extremepoints} (85.1\%) comes from the fact that the former is on PASCAL VOC 2012 segmentation validation, so \ours{} is trained on SBD + PASCAL train, whereas the latter is on a subset of PASCAL that overlaps with train (PASCAL$_{EXT}$), so \ours{} is only trained on COCO.